\documentclass[12pt]{article}

\usepackage{amsmath,amsthm,amsfonts}
\usepackage{graphicx,psfrag,epsf}
\usepackage{enumerate}
\usepackage{xcolor}
\usepackage{natbib}
\usepackage{multirow,multicol}
\usepackage{booktabs}
\usepackage{url} 
\usepackage[ruled,vlined]{algorithm2e}

\SetCommentSty{arxivcommentstyle}
\usepackage{soul}
\usepackage[mathscr]{euscript}
\usepackage{mathtools}
\usepackage{mathabx} %
\usepackage{makecell}
\usepackage{tabularx}
\usepackage{enumitem}
\usepackage{array} 
\usepackage[hidelinks]{hyperref} %
\graphicspath{{fig/}}

\newcommand{\blind}{1}

\def\spacingset#1{\renewcommand{\baselinestretch}{#1}\small\normalsize} 
\spacingset{1}

\newcolumntype{Y}{>{\raggedleft\arraybackslash}X}
\newcolumntype{Z}{>{\centering\arraybackslash}X}
\newcolumntype{P}[1]{>{\centering\arraybackslash}p{#1}}

\DeclareMathAlphabet{\pazocal}{OMS}{zplm}{m}{n} 
\DeclareFontFamily{OT1}{pzc}{}
\DeclareFontShape{OT1}{pzc}{m}{it}{<-> s * [1.10] pzcmi7t}{}

\newcommand{\bm}[1]{\boldsymbol{#1}}

\newcommand{\Fr}{{\mathrm{F}}}

\DeclareMathOperator*{\diag}{diag}

\definecolor{azure}{rgb}{0.0, 0.5, 1.0}

\numberwithin{equation}{section}

\newtheorem{theorem}{Theorem}[section]

\newtheorem{corollary}[theorem]{Corollary}

\newtheorem{proposition}[theorem]{Proposition}
\newtheorem{assumption}{Assumption}
\newtheorem{remark}{Remark}
\newtheorem{example}{Example}

\begin{document}

\if1\blind
{
  \title{\bf Enhancing Spectral Embedding through Robust and Flexible Knowledge Transfer in Electronic Health Records}
\author{
Feiqing Huang\textsuperscript{1},
Zongqi Xia\textsuperscript{2},
Rong Ma\textsuperscript{1,3,4}\thanks{Corresponding author: rongma@hsph.harvard.edu},
Tianxi Cai\textsuperscript{1,5}\thanks{Corresponding author: tcai@hsph.harvard.edu}
}

\date{
\textsuperscript{1} Department of Biostatistics, Harvard T.H. Chan School of Public Health \\
\textsuperscript{2} Department of Neurology, University of Pittsburgh
\\
\textsuperscript{3} Department of Data Science, Dana-Farber Cancer Institute \\
\textsuperscript{4} Broad Institute of MIT and Harvard \\ 
\textsuperscript{5} Department of Biomedical Informatics, Harvard Medical School
}
  \maketitle
} \fi

\if0\blind
{
  \bigskip
  \bigskip
  \bigskip
  \begin{center}
    {\LARGE\bf Enhancing Spectral Embedding through Robust and Flexible Knowledge Transfer in Electronic Health Records}
\end{center}
  \medskip
} \fi

\bigskip
\begin{abstract}
We propose a spectral-based, unsupervised representation learning framework to derive low-dimensional embeddings for clinical concepts and patients in rare disease cohorts from electronic health records, where data are high-dimensional but sample sizes are limited. To overcome this challenge, we incorporate a knowledge matrix extracted from a broader population that shares a partially overlapping subspace with the rare-disease cohort. Our method departs from existing approaches by relaxing restrictive one-to-one signal-alignment assumptions between the latent data matrix and knowledge matrix, allowing more flexible and realistic forms of structured sharing. We introduce a novel two-step spectral embedding procedure: first, we identify and remove irrelevant components from the knowledge matrix; then, we apply a projection-based method to separately recover shared and heterogeneous components. Simulations and an analysis of a real-world multiple sclerosis cohort show that the proposed method outperforms competing approaches, particularly in challenging scenarios where shared signals are weak and only partially aligned, as is common in rare-disease data.
\end{abstract}

\noindent%
\vfill

\newpage
\spacingset{1} %

\section{Introduction}

Rare diseases collectively affect millions of people worldwide, yet they are often difficult to study because individual conditions are uncommon and clinically heterogeneous. Multiple sclerosis (MS) is a representative example: although its global prevalence is only about 35.9 per 100,000 individuals \citep{walton2020rising}, it remains a major focus of clinical research because patients vary widely in symptoms, disease severity, and care patterns. This combination of low prevalence and substantial heterogeneity makes MS difficult to study systematically, as researchers must draw insight from limited cohorts while accounting for between-patient variation.

Electronic Health Records (EHRs) provide a rich and longitudinal repository of clinical data, capturing patient diagnoses, treatments, laboratory results, and clinical notes. This wealth of information has supported unsupervised representation learning methods that estimate low-dimensional embeddings without labels, enabling tasks such as patient profiling \citep{li2022graph, landi2020deep} and critical event identification \citep{weatherhead2022learning}. However, the inherently limited sample size of rare disease cohorts presents a major challenge for unsupervised representation learning methods, which typically require large datasets to learn robust representations. General biomedical knowledge graphs offer a potential solution by providing external information to enrich learning from scarce data. Yet, these graphs are often constructed with a focus on common diseases, which can lead to misalignment or irrelevant associations when directly applied to rare disease populations.

To state our problem, let $\bm{X}\in\mathbb{R}^{p\times n}$ denote the latent data matrix underlying EHR observations, where $n$ is the number of patients and $p$ is the number of clinical concepts (diagnoses, medications, procedures, laboratory tests, and related records). Because clinical concepts are often disease-specific and tend to co-occur in meaningful concept clusters and patient subgroups \citep{reza2022unsupervised}, we model $\bm{X}$ as low-rank. In practice, we observe a noisy version $\hat{\bm{X}}=\bm{X}+\bm{Z}$, where $\bm{Z}$ captures measurement noise and nuisance variation. Entries in $\hat{\bm{X}}$ can be binary indicators, code counts, or continuous laboratory values. Our goal is to learn low-dimensional embeddings for both clinical concepts and patients in rare disease cohorts, where $n$ can be much smaller than $p$. 
In many cases, prior knowledge about relationships among clinical concepts, derived from broader populations or external cohorts, is available and encoded as a reference concept embedding matrix $\bm{W} \in \mathbb{R}^{p \times q}$ with embedding dimension $q$ different from $p$; we refer to this as the knowledge matrix. We expect $\bm{W}$ to partially overlap with the latent concept structure in $\bm{X}$. Therefore, the central challenge is how to leverage $\bm{W}$ to improve learning of the low-rank structure of $\bm{X}$. 

\begin{figure}[t]
    \centering
    \includegraphics[width=.8\textwidth]{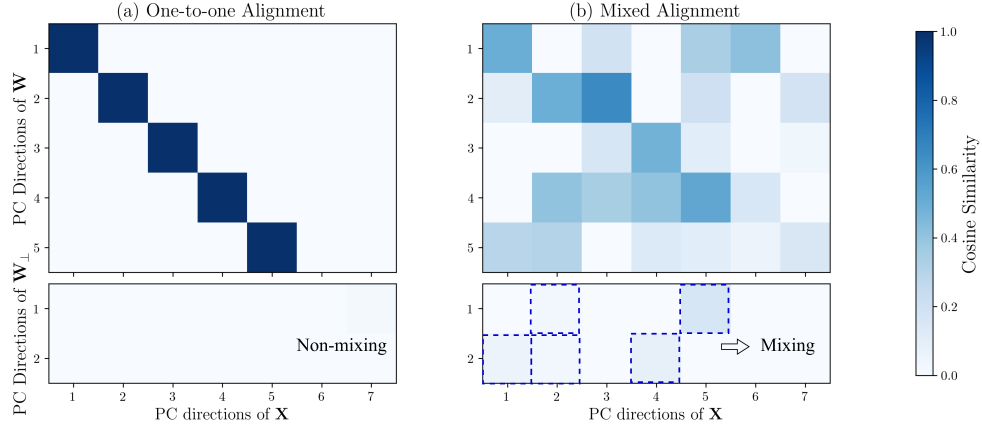}
    \caption{Cosine-similarity heatmaps between principal component (PC) directions of $\bm{W}$ and its orthogonal complement $\bm{W}_{\perp}$ (rows) and PC directions of $\bm{X}$ (columns).  
    (a) One-to-one alignment: each PC direction of $\bm{X}$ perfectly aligns with one PC direction of $\bm{W}$, with no alignment in $\bm{W}_{\perp}$. 
    (b) Mixed alignment: a PC direction of $\bm{X}$ can align with multiple directions across both $\bm{W}$ and $\bm{W}_{\perp}$, indicating cross-component mixing.}
    \label{fig:motivation}
\end{figure}

Numerous methods have been proposed to learn low-rank representations of $ \bm{X} $ with auxiliary knowledge. They can be broadly grouped into three families: spectral, manifold-based, and deep methods (Section~1.1). In the rare-disease setting, with limited sample size, high dimensionality, and strong interpretability requirements, we center our analysis on spectral methods and assume the latent data and auxiliary knowledge share partially overlapping subspaces. Within this setting, existing methods, including JIVE \citep{lock2013joint}, AJIVE \citep{feng2018angle}, CJIVE \citep{murden2022interpretive}, and others \citep{ma2024optimal}, face two key limitations. First, many existing spectral transfer methods assume \emph{one-to-one signal alignment}: each principal component (PC) direction in $ \bm{X} $ perfectly aligns with one PC direction in $ \bm{W} $; this yields the diagonal pattern in Figure~\ref{fig:motivation}(a). This assumption is often overly restrictive in real-world clinical data, where a PC direction of $\bm{X}$ aligns with multiple directions across $\bm{W}$ and $\bm{W}_{\perp}$ (mixed alignment; Figure~\ref{fig:motivation}(b)).

Second, existing methods typically rely on sufficiently strong shared signals between $ \bm{X} $ and $ \bm{W} $ and provide limited safeguards against \emph{negative transfer}, meaning that incorporating external knowledge leads to worse embedding estimation than using $ \hat{\bm{X}} $ alone (e.g., PCA). In rare-disease applications, clinically meaningful shared components are often subtle and can be overwhelmed by stronger heterogeneous variation (e.g., demographics or common comorbidities) \citep{zhao2025unveiling}, which makes transfer unstable. Although \cite{li2024knowledge} relaxes the one-to-one alignment assumption
through subspace-distance minimization, it does not explicitly address this weak-shared-signal negative-transfer regime. The main contributions of this work can be summarized as follows:

\begin{figure}[t]
    \centering
    \includegraphics[width=1.\textwidth]{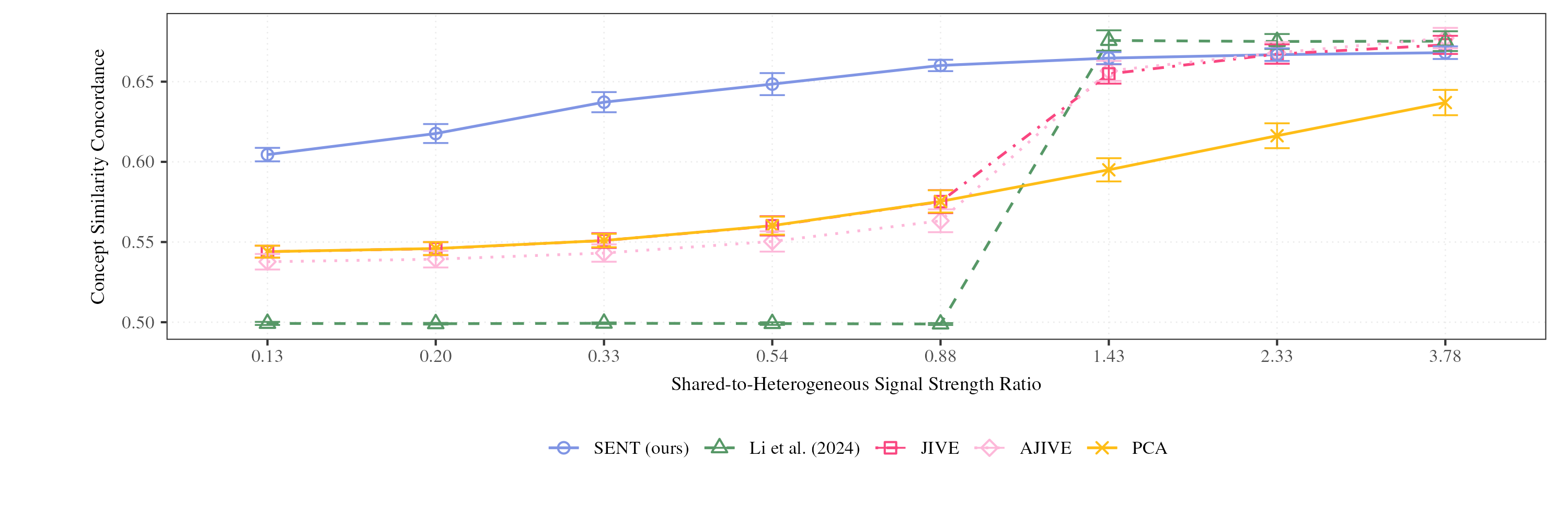}
    \caption{Performance comparison across simulated datasets with varying shared-to-heterogeneous signal strength ratios. The y-axis shows the concordance index between the estimated concept similarity structure, computed from the estimated concept embeddings, and the ground truth, with higher values indicating better recovery. When the shared signal is weak relative to the heterogeneous signal (ratio$<$1), \cite{li2024knowledge} and AJIVE underperform PCA, indicating negative transfer relative to a non-transfer baseline.
    }
    \label{fig:motivation_algorithm}
\end{figure}

  \begin{itemize}
  \item \textbf{Flexible transfer beyond one-to-one alignment.}
  We relax the restrictive one-to-one signal-alignment assumption underlying many existing knowledge-transfer methods and develop a general subspace-based transfer framework. The framework is built on subspace-distance characterization and a newly introduced nonseparability parameter, which jointly capture mixed cross-component correspondence between $ \bm{X} $ and $ \bm{W} $ and support a more refined theoretical analysis.

  \item \textbf{A two-step design to mitigate negative transfer under weak shared signal.}
  We introduce a two-step estimator tailored to rare-disease settings where shared signal is weak relative to strong heterogeneous variation: Step I preprocesses the raw knowledge matrix to remove non-transferable directions, and Step II estimates shared and heterogeneous components through projection onto $ \mathrm{span}(\bm{W}) $ and $ \mathrm{span}(\bm{W}_{\perp}) $, respectively. This design improves recovery of weak shared structure and provides robustness against negative transfer where direct transfer is unstable (Figure~\ref{fig:motivation_algorithm}).

  \item \textbf{Non-asymptotic guarantees for concept and patient embeddings.}
  We provide rigorous theory for both preprocessing and estimation, including consistency of the preprocessing step and non-asymptotic error bounds for concept and patient embeddings. In particular, Theorem~\ref{thm:factor_statistical_ub} shows that informative concept-level transfer improves not only concept embedding but also patient embedding estimation.
  \end{itemize}

To establish these results, we develop technical tools that align with the three contributions above. We formalize mixed sharing through subspace-distance and nonseparability quantities, characterize when preprocessing avoids negative transfer, and derive non-asymptotic guarantees for both concept and patient embeddings. In particular, we establish deterministic upper and lower bounds that track strong and weak signal components individually, rather than analyzing only their aggregate linear span as in prior work \citep{li2024knowledge,ma2024optimal}. This yields a sharper and more interpretable decomposition of subspace estimation error into deterministic bias and stochastic fluctuation. See Supplementary Section~S1 for details.

The remainder of the paper is organized as follows. Section~1.1 reviews related work. Section~2 introduces the model setup and geometric quantities for knowledge transfer. Section~3 presents the proposed two-step SENT procedure, including knowledge preprocessing and embedding estimation. Section~4 establishes deterministic and statistical guarantees for concept and patient embeddings. Sections~5 and~6 report simulation and real-data results, respectively. Section~7 concludes with a discussion.

\subsection{Related work}

Unsupervised representation learning methods can be broadly grouped into three families: spectral methods (e.g., PCA and factor models) \citep{bai2008large,fan2021robust}, manifold-based methods (e.g., t-SNE and UMAP) \citep{maaten2008visualizing,mcinnes2018umap}, and deep representation methods (e.g., generative and contrastive approaches) \citep{mansour2021unsupervised,chen2020simple}. In this paper, our focus is the rare-disease regime with limited sample size, high dimensionality, and a strong need for interpretability. In such settings, manifold and deep methods can be less stable and harder to interpret geometrically, especially when sample size is small \citep{arvanitidis2017latent,chennuru2018measures}. We therefore focus on spectral approaches, which provide a transparent and statistically tractable foundation for embedding estimation \citep{allen2023interpretable}.

Given this spectral focus, methods that incorporate auxiliary knowledge can be grouped into four classes: linear subspace alignment \citep{lock2013joint,feng2018angle,ma2024optimal}, manifold-geometry transfer \citep{ding2025kernel,landa2024entropic}, distributional alignment \citep{pan2010domain,david2010impossibility}, and deep transfer methods such as adversarial or discrepancy-based alignment \citep{jiao2024deep}. These approaches provide useful transfer mechanisms when geometric and distributional assumptions are matched across domains. However, in clinical transfer settings with substantial cohort heterogeneity, manifold correspondence may be unstable, distributional alignment can be sensitive to covariate shift, and deep transfer methods often require larger sample sizes in both source and target domains. For these reasons, subspace-based spectral transfer methods are the most directly relevant comparison class for our problem.

{\bf Notation.} Scalars are denoted by lower-case letters, vectors by bold lower-case letters, and matrices by bold upper-case letters. Universal constants are denoted by $C,c,\ldots$ and may vary from line to line. For $a,b\in\mathbb{R}$, let $a\wedge b=\min\{a,b\}$. For sequences $\{a_n\}$ and $\{b_n\}$, write $a_n\lesssim b_n$ (or $a_n\gtrsim b_n$) if $a_n\le Cb_n$ (or $a_n\ge Cb_n$) for some universal constant $C>0$, and $a_n\asymp b_n$ if both hold. For deterministic sequences, $a_n=O(b_n)$ means $|a_n|\le Cb_n$ eventually. For random sequences, $X_n=O_p(a_n)$ means $X_n/a_n$ is bounded in probability. For a matrix $\bm{X}$, let $\|\bm{X}\|$ and $\|\bm{X}\|_{\mathrm{F}}$ denote the operator and Frobenius norms. Let $\mathcal{O}_{p,q}=\{\bm{X}\in\mathbb{R}^{p\times q}:\bm{X}^\top\bm{X}=\bm{I}_q\}$, and write $\mathcal{O}_p$ when $p=q$. Supplementary Table~S1 gives a complete notation table.

\section{Matrix Denoising Model with Subspace Knowledge}
\subsection{Model setup}\label{sec:model_setup}

In this subsection, we introduce the matrix denoising model, define alignment between auxiliary knowledge and the latent data subspace via principal angles, and use this alignment to decompose the latent subspace into a shared (knowledge-driven) component and a heterogeneous (purely data-driven) component.

Suppose we observe $\hat{\bm{X}}\in\mathbb{R}^{p\times n}$, generated from a rank-$r$ latent data matrix $\bm{X}$ plus perturbation $\bm{Z}$:
\begin{align}\label{eq:Xhat}
\hat{\bm{X}} = \bm{X}+\bm{Z}, \quad \text{where}\quad \bm{X} \overset{\mathrm{SVD}}{=} \bm{U}\bm{\Lambda}^{1/2}\bm{V}^{\top}=\bm{U}\bm{F}^{\top}.
\end{align}
Here $ \bm{U}\in\mathcal{O}_{p,r} $ and $ \bm{V}\in\mathcal{O}_{n,r} $ are the left and right singular vector matrices, and $ \bm{\Lambda}=\diag(\lambda_1, \ldots,\lambda_r) $ with $ \lambda_1\ge\cdots\ge\lambda_r>0 $. For EHR applications, we interpret $ \bm{U} $ as latent concept embeddings and $ \bm{F}=\bm{V}\bm{\Lambda}^{1/2} $ as patient embeddings.

Let $ \bm{W}\in\mathcal{O}_{p,r_W} $ be a transferable knowledge matrix, where $ r_W\le r $. The integer $ r_W $ is the transferable knowledge rank (estimated in Section~3), and in Section~2 we treat $ \bm{W} $ as the ideal informative knowledge matrix. To characterize how $ \bm{W} $ aligns with the latent subspace $\mathrm{span}(\bm{U}) $, consider the SVD of $ \bm{U}^{\top}\bm{W} $:
\begin{equation}\label{eq:UTW_svd}
    \bm{U}^{\top}\bm{W} =
    [\underset{r\times r_W}{\bm{Q}_1} \ \underset{r\times (r-r_W)}{\bm{Q}_2}]
    \begin{bmatrix}
    \diag(\sigma_1,\ldots,\sigma_{r_W})\\
    \bm{0}
    \end{bmatrix}
    \bm{O}^{\top},
\end{equation}
where $[\bm{Q}_1 \ \bm{Q}_2]\in\mathcal{O}_r$, $\bm{O}\in\mathcal{O}_{r_W}$, and $1\ge \sigma_1\ge\cdots\ge \sigma_{r_W}>0$.
Define the principal angles by
\begin{align}\label{eq:def_angle}
    \theta_j := \arccos(\sigma_j), \quad 1 \leq j \leq r_W.
\end{align}
Smaller $ \theta_j $ indicates stronger alignment between the knowledge subspace and the latent data subspace. The matrix $ \bm{Q}_1 $ captures directions in $\mathrm{span}(\bm{U}) $ shared with $ \bm{W} $, while $ \bm{Q}_2 $ captures heterogeneous directions orthogonal to $ \bm{W} $.

The angles $ \theta_j $ are uniquely defined. For fixed $ \bm{U} $, $ \bm{Q}_2 $ is identifiable up to right multiplication by an orthogonal matrix in $\mathcal{O}_{r-r_W} $. The identifiability of $ \bm{Q}_1 $ depends on multiplicities of $ \sigma_j $: it is unique up to sign when the $ \sigma_j $ are distinct, and up to blockwise orthogonal rotation within groups of repeated singular values otherwise.

This decomposition gives the shared-versus-heterogeneous split used by our estimator in Section~\ref{sec:estimation} and by the nonseparability analysis in Section~\ref{sec:nonseparability}.

\subsection{Subspace distance, nonseparability, and flexible sharing}\label{sec:nonseparability}

To formalize flexible knowledge transfer and support later theory, we introduce subspace-distance and nonseparability quantities. For orthonormal basis matrices $ \bm{A}\in\mathcal{O}_{p,q} $ and $ \bm{B}\in\mathcal{O}_{p,k} $ with principal angles $ \phi_1\le\cdots\le\phi_{q\wedge k} $ and $q\wedge k=\min(q, k)$, define
  \begin{align}\label{eq:subspace_notation}
      \|\sin\Theta(\bm{A},\bm{B})\|:=\max_{1\le j\le q\wedge k}\sin\phi_j,\qquad
  \|\sin\Theta(\bm{A},\bm{B})\|_{\Fr}:=\Big(\sum_{j=1}^{q\wedge k}\sin^2\phi_j\Big)^{1/2}.
  \end{align}
Applying this to $ \bm{U} $ and $ \bm{W} $, we define
  \begin{align}\label{eq:def_subspace_distance}
          \gamma := \|\sin\Theta(\bm{U},\bm{W})\| = \sin\theta_{r_W}, \qquad
          \gamma_{\Fr} := \|\sin\Theta(\bm{U},\bm{W})\|_{\Fr}= \Big(\sum_{j=1}^{r_W}\sin^2\theta_j\Big)^{1/2},
  \end{align}
where $ \theta_j $ are the principal angles in \eqref{eq:def_angle} and $ \theta_{r_W} $ is the largest.
  
Let $ \mathcal{U}=\text{span}(\bm{U}) $ and $ \mathcal{W}=\text{span}(\bm{W}) $. Since $ \bm{X}\bm{X}^\top=\bm{U}\bm{\Lambda}\bm{U}^\top $, rotating the column vectors of $ \bm{U} $ by $ [\bm{Q}_1\ \bm{Q}_2] $ from \eqref{eq:UTW_svd} gives
  \begin{align*}
      \bm{X}\bm{X}^{\top}
      =
      \underbrace{[\bm{U}\bm{Q}_1\ \bm{U}\bm{Q}_2]}_{\text{rotated eigenbasis}}
      \underbrace{\begin{bmatrix}
          \bm{Q}_1^{\top}\bm{\Lambda}\bm{Q}_1 & \bm{Q}_1^{\top}\bm{\Lambda}\bm{Q}_2\\
          (\bm{Q}_1^{\top}\bm{\Lambda}\bm{Q}_2)^{\top} & \bm{Q}_2^{\top}\bm{\Lambda}\bm{Q}_2
      \end{bmatrix}}_{\text{rotated eigenvalue blocks}}
      [\bm{U}\bm{Q}_1\ \bm{U}\bm{Q}_2]^{\top}.
  \end{align*}
Geometrically, this decomposition splits $ \mathcal{U} $ into two orthogonal components: $ \text{span}(\bm{U}\bm{Q}_1) $, which overlaps with $ \mathcal{W}$, and $ \text{span}(\bm{U}\bm{Q}_2) $, which is orthogonal to $ \mathcal{W} $.
The middle block matrix shows how signal is allocated across the two components: the diagonal blocks $\bm{Q}_1^\top\bm{\Lambda}\bm{Q}_1 $ and $ \bm{Q}_2^\top\bm{\Lambda}\bm{Q}_2 $ describe signal variation within each component, while the off-diagonal block $ \bm{Q}_1^\top\bm{\Lambda}\bm{Q}_2 $ measures how strongly the two components are mixed.

If the two components are not mixed, $ \bm{Q}_1^\top\bm{\Lambda}\bm{Q}_2=\bm{0} $ and we then have \textit{perfect separation}. The one-to-one alignment setting, which underlies many multi-matrix learning approaches such as JIVE \citep{lock2013joint}, is perfectly separated (see Example~\ref{ex:JIVE}).

However, this setting is often unrealistic. In practice, $\mathcal{U}$ typically overlaps with both $\mathcal{W}$ and its orthogonal subspace, leading to the mixed alignment setting where $ \bm{Q}_1^{\top}\bm{\Lambda}\bm{Q}_2\neq \bm{0} $ (see Example~\ref{ex:our}). To quantify this, we define the signal \textit{nonseparability} parameter
  \begin{equation}\label{eq:def_delta}
      \delta := \|\bm{Q}_1^{\top}\bm{\Lambda}\bm{Q}_2\|.
  \end{equation}
Larger $ \delta $ indicates stronger mixing between the shared and heterogeneous components. When all eigenvalues are identical, $ \delta=0 $ because $\bm{X} $ can be represented using $ [\bm{U}\bm{Q}_1,\bm{U}\bm{Q}_2] $ without changing $ \bm{\Lambda} $, which is effectively another perfectly separated case. When eigenvalues differ, $ \delta $ is generally nonzero and captures the extent of cross-component mixing.

\begin{example}[One-to-one alignment]\label{ex:JIVE}
  Consider $ \bm{W}=[\bm{w}_1,\ldots,\bm{w}_{r_W}]\in\mathcal{O}_{p,r_W} $ and $ \bm{X}=\bm{U}\bm{\Lambda}^{1/2} $, where $ \bm{U}=[\bm{u}_1,\ldots,\bm{u}_r]\in\mathcal{O}_{p,r} $ and $ \bm{\Lambda}=\textup{diag}(\lambda_1,\dots,\lambda_r) $ with distinct eigenvalues. Suppose each $ \bm{w}_j $ forms a small angle $ \theta_j $ with $ \bm{u}_j $ and is orthogonal to all $ \bm{u}_i $ for $ i\neq j $. Then
  $\bm{Q}_1 = [\bm{I}_{r_W}, \bm{0}_{r_W,r-r_W}]^{\top}$ and $\bm{Q}_2 = [\bm{0}_{r-r_W,r_W}, \bm{O}_{r-r_W}]^{\top}$,
  where $ \bm{O}_{r-r_W}\in\mathcal{O}_{r-r_W} $. Hence $ \bm{Q}_1^\top\bm{\Lambda}\bm{Q}_2=\bm{0} $, so $ \delta=0 $.
\end{example}

\begin{example}[Mixed alignment]\label{ex:our}
Consider $ r=2 $ and $ r_W=1 $. Write $ \bm{W}=[\bm{w}] $ with $ \|\bm{w}\|=1 $, and let $\bm{X}=\bm{U}\bm{\Lambda}^{1/2} $, where $ \bm{U}=[\bm{u}_1,\bm{u}_2]\in\mathcal{O}_{p,2} $ and $ \bm{\Lambda}=\textup{diag}(\lambda_1,\lambda_2) $ with $ 0<\lambda_1<\lambda_2$. Since $r_W=r-r_W=1$, write $\bm{Q}_1=[\bm{q}_1]$ and $\bm{Q}_2=[\bm{q}_2]$ for unit vectors $\bm{q}_1,\bm{q}_2\in\mathbb{R}^2$. Let $\bm{u}_{\perp}$ be a unit vector orthogonal to $\mathrm{span}(\bm{U})$, and set $ \bm{w} = \sqrt{1-\gamma^2}\big(\gamma\bm{u}_1+\sqrt{1-\gamma^2}\bm{u}_2\big) + \gamma \bm{u}_{\perp} $ for $ 0<\gamma<0.5 $. Then $ \|\sin\Theta(\bm{U},\bm{W})\|=\gamma $. Moreover, the normalized projection of $\bm{w}$ onto $\mathrm{span}(\bm{U})$ is $\gamma\bm{u}_1+\sqrt{1-\gamma^2}\bm{u}_2$, so $ \bm{q}_1 = [\gamma,\sqrt{1-\gamma^2}]^{\top} $, $ \bm{q}_2 = [-\sqrt{1-\gamma^2},\gamma]^{\top} $, and hence $ \delta = \|\bm{Q}_1^{\top}\bm{\Lambda}\bm{Q}_2\| = |\bm{q}_1^{\top}\bm{\Lambda}\bm{q}_2| = \gamma\sqrt{1-\gamma^2}(\lambda_2-\lambda_1)>0 $.
\end{example}

As shown in Section~4, the subspace-distance quantities $ \gamma $ and $ \gamma_{\Fr} $, together with the nonseparability parameter $ \delta $, are central to characterizing the relationship between the latent data and auxiliary knowledge. They appear explicitly in both upper and lower bounds for embedding estimation error (see Theorem~4.1).

Section~2 provides the geometric setup under a transferable knowledge matrix $ \bm{W} $. In practice, we start from a raw matrix $ \bm{W}_0 $, which may combine transferable and non-transferable directions. Section~3 introduces a preprocessing step to construct a transferable matrix from $\bm{W}_0 $ and gives its sample algorithm. Section~4 then develops theory using Section~2 quantities for the transferable knowledge matrix $ \bm{W}$ constructed in Section~3.

\section{Spectral Embedding with Knowledge Transfer} \label{sec:estimation}

We present {\bf S}pectral {\bf E}mbedding with k{\bf N}owledge {\bf T}ransfer (SENT) as a two-step procedure for estimating concept and patient embeddings. For simplicity, we assume the true rank $r$ of $\bm{X}$ is known; in practice, $r$ can be consistently estimated using existing rank-selection methods \citep{choi2017selecting,passemier2014estimation,ke2023estimation,donoho2023screenot}.

The practical inputs in this section are the observed matrix $\hat{\bm{X}}$ and a raw knowledge matrix $\bm{W}_0$. Because $\bm{W}_0$ may contain both transferable and non-transferable directions, using it directly can induce negative transfer. Step I estimates a transferable rank and a transferable knowledge matrix from $\bm{W}_0$; Step II uses this knowledge matrix to estimate embeddings. The corresponding sample estimators are given in Algorithms~\ref{alg:data_preprocessing} and~\ref{alg:estimation}.

Without auxiliary knowledge, the standard non-transfer baseline estimator of $ \bm{U} $ is given by the top-$r$ left singular vectors of $ \hat{\bm{X}} $, denoted by $ \hat{\bm{U}}^{(0)} $ \citep{bai2002determining,cai2018rate_optimal}.
Its consistency requires regularity conditions on the perturbation term $ \bm{Z} $. Under Assumption~\ref{assum:Z}, there exist sequences $\varepsilon_0=\varepsilon_0(n,p)\to 0 $ and $ \eta_{\varepsilon_0}\to 0 $ such that $\mathbb{P}\!\left(\|\sin\Theta(\hat{\bm{U}}^{(0)},\bm{U})\|_{\Fr}\le \varepsilon_0\right)\ge 1-\eta_{\varepsilon_0}.$
Transferable external knowledge can improve this baseline error, whereas irrelevant or adversarial knowledge can worsen it (negative transfer) \citep{li2024knowledge,zhang2022survey}. This motivates the preprocessing step in Step~I.

\subsection{Step I: Knowledge preprocessing}

Section~2 introduced the population notion of transferable knowledge. In Step~I, we map a raw knowledge matrix $ \bm{W}_0\in\mathbb{R}^{p\times d} $ to a transferable matrix $ \bm{W} $ and its rank $ r_W $, where $d$ is the column dimension of $ \bm{W}_0 $ with $d \leq p$.

We compute the SVD of $ \bm{W}_0 $ and denote its left singular vectors by $ \tilde{\bm{W}}\in\mathcal{O}_{p,d} $, that is, the knowledge basis matrix. Let $[\bm{o}_1,\ldots,\bm{o}_d]\in\mathcal{O}_d$ be the right singular vectors of $ \bm{U}^{\top}\tilde{\bm{W}} $, and define $ \bm{O}_{1,k}=[\bm{o}_1,\ldots,\bm{o}_k]\in\mathcal{O}_{d,k} $ for $0\le k\le
r\wedge d$.
Then define candidate knowledge matrices
$\bm{W}_k:=\tilde{\bm{W}}\bm{O}_{1,k}$.
Each is a rotated and truncated version of the knowledge basis $\tilde{\bm{W}}$.
While larger $k$ retains more directions from the knowledge basis, it may also introduce non-transferable ones.
Hence, a candidate is transferable if its subspace distance to $ \bm{U} $ is at most the error order $ \varepsilon_0 $ for the non-transfer baseline estimator $ \hat{\bm{U}}^{(0)} $. Specifically, let %
\begin{equation}\label{eq:def_Wset}
    \mathbb{W}_{U,k}
    =
    \left\{
    \bm{A}\in\mathcal{O}_{p,k}
    :\ \exists c_W>0\ \text{(independent of $n,p$) s.t.}\
    \|\sin\Theta(\bm{A},\bm{U})\|_{\Fr}\le c_W\varepsilon_0
    \right\},
\end{equation}
for each $k\ge1$, and set $ \mathbb{W}_{U,0}=\emptyset $. We call $ \bm{W}_k $ a transferable candidate if $ \bm{W}_k\in\mathbb{W}_{U,k} $.

Among transferable candidates, we choose the largest rank as the \textit{transferable rank}
\begin{align}\label{eq:r_construct}
    r_W=\max\!\left(\{0\}\cup\left\{1\le k\le r\wedge d:\ \bm{W}_k\in\mathbb{W}_{U,k}\right\}\right).
\end{align}
The \textit{transferable knowledge matrix} is then set to the corresponding candidate,
\begin{equation}\label{eq:W_construct}
\bm{W}=\bm{W}_{r_W}\in\mathbb{W}_{U,r_W}\quad\text{if } r_W>0,\qquad
\bm{W}=\emptyset\quad\text{if } r_W=0.
\end{equation}
If $r_W=0$, no direction extracted from $ \bm{W}_0 $ satisfies the transferability criterion in \eqref{eq:def_Wset}.

  \begin{algorithm}[t]
        \caption{Knowledge preprocessing}
        \label{alg:data_preprocessing}
        \textbf{Input:} The baseline left singular matrix estimator $\hat{\bm{U}}^{(0)}\in\mathcal{O}_{p,r}$, its error level $\varepsilon_0$, a fixed constant $c_W>0$, and the raw knowledge matrix $\bm{W}_0\in\mathbb{R}^{p\times d}$. \\\vspace{2mm}
        \tcp{Build an orthogonal knowledge basis matrix $\tilde{\bm{W}}$ from the raw knowledge matrix $\bm{W}_0$.}
        1. \hspace{2mm}Perform SVD on $\bm{W}_0$ and let $\tilde{\bm{W}}\in\mathcal{O}_{p,d}$ be its left singular vectors.\\
        \tcp{Construct nested candidate matrices $\hat{\bm{W}}_k$ using $\hat{\bm{U}}^{(0)}$. Each candidate $\hat{\bm{W}}_k$ is a rotated and truncated version of the knowledge basis matrix $\tilde{\bm{W}}$.}
        2. \hspace{2mm}Perform SVD on $\hat{\bm{U}}^{(0)\top}\tilde{\bm{W}}$ to obtain right singular vectors $\hat{\bm{o}}_k^{(0)}\in\mathbb{R}^{d}$ for $1\le k\le
  r\wedge d$. Set
        \[
        \hat{\bm{O}}_{1,k}^{(0)}=[\hat{\bm{o}}_1^{(0)},\ldots,\hat{\bm{o}}_k^{(0)}],\qquad
        \hat{\bm{W}}_k=\tilde{\bm{W}}\hat{\bm{O}}_{1,k}^{(0)}.
        \]
        \tcp{Select the largest transferable rank and the transferable knowledge matrix.}
        3. \hspace{2mm}Define
        \begin{align}\label{eq:def_hatr_theta}
        \hat{r}_W :=\max\!\left(\{0\}\cup\{1\le k\le r\wedge d:\ \|\sin\Theta(\hat{\bm{W}}_k,\hat{\bm{U}}^{(0)})\|_{\Fr}\le c_W\varepsilon_0\}\right).
        \end{align}
        And set $\hat{\bm{W}}=\hat{\bm{W}}_{\hat r_W}$.\\
        \textbf{Return} $(\hat r_W,\hat{\bm{W}})$.\\\vspace{2mm}
  \end{algorithm}

Algorithm~\ref{alg:data_preprocessing} is the sample analogue of \eqref{eq:r_construct}--\eqref{eq:W_construct}: it replaces unknown $ \bm{U} $ with $\hat{\bm{U}}^{(0)} $ and returns $(\hat r_W,\hat{\bm{W}})$. Since $ \varepsilon_0 $ is unknown, in practice we calibrate it either from the analytic rate in \cite{cai2018rate_optimal} or from the empirical heuristics described in Supplementary Section~S2. If $\hat r_W=r\wedge d$, then $\hat{\bm{W}}=\tilde{\bm{W}}$; if $\hat r_W=0$, then $\hat{\bm{W}}=\emptyset$ (no transferable knowledge). The next proposition establishes consistency of this preprocessing step, focusing on the case $0<r_W<r\wedge d$.

\begin{proposition}[Asymptotic consistency of knowledge preprocessing]\label{prop:knowledge_matrix}
  Let $c_W>0$ be a sufficiently large fixed constant. Suppose there exists a sequence $\varepsilon_0=\varepsilon_0(n,p)\to0$
  such that $\lim_{n,p\to\infty}\mathbb{P}\!\left(\|\sin\Theta(\hat{\bm{U}}^{(0)},\bm{U})\|_{\Fr}\le\varepsilon_0\right)=1.$
  Let $r_W=r_W(\varepsilon_0)$ be defined in \eqref{eq:r_construct}, and assume $0<r_W<r\wedge d$. Define $\hat r_W$ and $
  \hat{\bm{W}}_k$ as in Algorithm~\ref{alg:data_preprocessing}. If
  $\liminf_{(n,p)\to\infty}\big(\sigma_{r_W}-\sigma_{r_W+1}\big)\ge c_{\mathrm{gap}}>0,$
  where $\sigma_k$ is the $k$-th singular value of $\bm{U}^{\top}\tilde{\bm{W}}$ and $c_{\mathrm{gap}}$ is a fixed constant, then
  \[
  \lim_{n,p\to\infty}\mathbb{P}\!\left(\begin{aligned}
  \|\sin\Theta(\hat{\bm{W}}_{r_W},\hat{\bm{U}}^{(0)})\|_{\Fr}&\le c_W\varepsilon_0,\\
  \|\sin\Theta(\hat{\bm{W}}_{r_W+1},\hat{\bm{U}}^{(0)})\|_{\Fr}&\ge \sqrt{1-(1-c_{\mathrm{gap}}/4)^2}
  \end{aligned}\right)=1.
  \]
  Consequently,
  \[
  \lim_{n,p\to\infty}\mathbb{P}\!\left(\hat r_W=r_W,\ \hat{\bm{W}}\in\mathbb{W}_{U,r_W}\right)=1.
  \]
  \end{proposition}

\subsection{Step II: Embedding estimation with knowledge transfer}

Step~I produces a transferable knowledge matrix and its rank. In this subsection, we describe Step~II under the corresponding population setup: the transferable matrix is denoted by $ \bm{W} $ with rank $r_W$ (defined in \eqref{eq:r_construct} and \eqref{eq:W_construct}). Let $ \bm{W}_\perp\in\mathbb{R}^{p\times(p-r_W)} $ be an orthonormal basis for the orthogonal complement of $ \bm{W} $. 
Algorithm~\ref{alg:estimation} uses these quantities to estimate the concept embeddings $\bm{U}$ and patient embeddings $\bm{F}$. In practice, $(r_W,\bm{W})$ are replaced by the Step~I outputs $(\hat r_W,\hat{\bm{W}})$. 
  
  \begin{algorithm}[h!]
  \caption{Embedding estimation with knowledge transfer}
  \label{alg:estimation}
  \KwIn{Data matrix $\hat{\bm{X}} \in \mathbb{R}^{p \times n}$, filtered knowledge matrix $\bm{W} \in \mathbb{R}^{p \times r_W}$, its orthogonal complement $\bm{W}_\perp$, embedding dimension $r$, knowledge rank $r_W$}
  \KwOut{Estimated subspace $\hat{\bm{U}} \in \mathbb{R}^{p \times r}$ and patient embeddings $\hat{\bm{F}} \in \mathbb{R}^{n \times r}$}
  \tcp{Knowledge-driven block: project to $\mathrm{span}(\bm{W})$ to extract shared directions.}
  1. Compute the top $r_W$ eigenvectors of $\bm{W}^\top\hat{\bm{X}}\hat{\bm{X}}^\top\bm{W}$, denoted $\hat{\bm{\Omega}}_w$\;
  2. Set $\hat{\bm{U}}_w=\bm{W}\hat{\bm{\Omega}}_w$\;
  \tcp{Data-driven block: project to $\mathrm{span}(\bm{W}_\perp)$ to recover heterogeneous directions.}
  \If{$r_W<r$}{
  3. Compute the top $r-r_W$ eigenvectors of $\bm{W}_\perp^\top\hat{\bm{X}}\hat{\bm{X}}^\top\bm{W}_\perp$, denoted $\hat{\bm{\Omega}}_{w^\perp}$\;
  4. Set $\hat{\bm{U}}_{w^\perp}=\bm{W}_\perp\hat{\bm{\Omega}}_{w^\perp}$\;
  }
  \tcp{Combine both blocks for full estimated concept and patient embeddings.}
  5. Construct
  \[
  \hat{\bm{U}}=
  \begin{cases}
  \hat{\bm{U}}_w, & r_W=r,\\
  [\hat{\bm{U}}_w\ \hat{\bm{U}}_{w^\perp}], & 0<r_W<r,\\
  \hat{\bm{U}}^{(0)}, & r_W=0.
  \end{cases}
  \]
  Set $\hat{\bm{F}}=\hat{\bm{X}}^\top\hat{\bm{U}}$.
  \end{algorithm}

Algorithm~\ref{alg:estimation} combines a knowledge-driven block and a data-driven block. The eigenvectors of $ \bm{W}^{\top}\hat{\bm{X}}\hat{\bm{X}}
^{\top}\bm{W} $ recover shared directions within the knowledge subspace, while the eigenvectors of $\bm{W}_{\perp}^{\top}\hat{\bm{X}}\hat{\bm{X}}^{\top}\bm{W}_{\perp} $ recover heterogeneous directions from its orthogonal complement. The estimator adapts to the informativeness of knowledge: it reduces to classical SVD
when $r_W=0$, uses only knowledge-guided directions when $r_W=r$, and combines both blocks when $0<r_W<r$. In practice, $ \hat{\bm{U}} $ and $ \hat{\bm{F}} $ are identifiable only up to an orthogonal transformation, which is sufficient for downstream analyses on concept similarity from concept embeddings and supervised prediction from patient embeddings.

\begin{remark}
  Given a transferable knowledge matrix $ \bm{W} $, Algorithm~1 of \cite{li2024knowledge} can also be applied for transfer to $ \bm{X} $. In our notation, their method corresponds to the case when $0<r_W<r$ and the knowledge side is noisy. By contrast, our
  framework first estimates $r_W$ and a ``noise-less'' knowledge matrix via Algorithm~\ref{alg:data_preprocessing} and then handles the full range of cases, including $r_W = r$ and $r_W = 0$.
\end{remark}

\begin{remark}
  To estimate the shared component, JIVE and AJIVE use a stacked-SVD step (see equation~(6) in \cite{feng2018angle}). That construction is not designed to handle mixed or rotationally misaligned signal components between knowledge and data, whereas Algorithm~\ref{alg:estimation} accommodates this through separate projections onto $ \mathrm{span}(\bm{W}) $ and $ \mathrm{span}(\bm{W}_\perp) $.
\end{remark}

\section{Theoretical Results} \label{sec:theory}

\subsection{Deterministic bounds for subspace analysis} \label{subsec:deterministic}

In Section~4, assume $1\le r_W<r$ and $p\ge r+r_W$. All quantities are defined with respect to the transferable knowledge matrix $ \bm{W}\in\mathbb{R}^{p\times r_W} $ in \eqref{eq:W_construct} with rank $r_W$, rather than the raw knowledge matrix $ \bm{W}_0 $. Proposition~\ref{prop:knowledge_matrix} shows that the sample estimator $\hat{\bm{W}}$ from Step~I consistently recovers $\bm{W}$.
In this subsection, we study deterministic error bounds for $\hat{\bm{U}}$ from Algorithm~\ref{alg:estimation}; full technical forms are given in Theorems~S.1 and~S.2 in the Supplementary.

In \eqref{eq:UTW_svd}, $ \bm{Q}_1 $ and $ \bm{Q}_2 $ split $ \mathrm{span}(\bm{U}) $ into shared and heterogeneous directions. Recall that $ \gamma=\|\sin\Theta(\bm{U},\bm{W})\| $ and $ \gamma_{\Fr}=\|\sin\Theta(\bm{U},\bm{W})\|_{\Fr} $ measure subspace distance (see \eqref{eq:def_subspace_distance}), while $
\delta=\|\bm{Q}_1^{\top}\bm{\Lambda}\bm{Q}_2\| $ measures cross-component mixing (see \eqref{eq:def_delta}). Define
\[
D:=\lambda_{Q_2,\min}-\gamma^2\lambda_{Q_1,\max},
\qquad
e_\delta:=\frac{\delta\gamma_{\Fr}}{D},
\]
where $ \lambda_{Q_1,\max} $ and $ \lambda_{Q_2,\min} $ are the largest and smallest eigenvalues of $ \bm{Q}_1^{\top}\bm{\Lambda}\bm{Q}_1 $ and $ \bm{Q}
_2^{\top}\bm{\Lambda}\bm{Q}_2 $, respectively. Here, $D$ measures the gap between heterogeneous and shared signals, and $e_\delta$ is the resulting deterministic mixing bias.
Let $ e_{\text{rand}} $ denote the stochastic error from finite-sample noise after projection onto $ \mathrm{span}(\bm{W}_{\perp}) $, as defined in
  Theorem~S.1.

\begin{theorem}[Deterministic subspace estimation bounds (informal)]\label{thm:informal_deterministic}
  Under the conditions of Theorems~S.1 and~S.2 in the Supplementary, if $ D>2\delta\gamma $, then
  \begin{align}
  \|\sin\Theta(\hat{\bm{U}}, \bm{U})\|_{\Fr}
  \le
  \min\!\left(\gamma_{\Fr} + \frac{4}{3}e_{\delta} + e_{\text{rand}},\sqrt{r}\right). \label{ub.eq}
  \end{align}
  In addition, for fixed $\bm W$, the corresponding minimax lower bound satisfies
  \begin{align}
  \inf_{\tilde{\bm{U}}}
  \sup_{\vartheta\in\mathcal{G}_{\bm W}(\gamma_{\Fr},e_{\delta},e_{\text{rand}})}
  \|\sin\Theta(\tilde{\bm{U}},\bm{U})\|_{\Fr}
  \gtrsim
  \max\!\left\{\gamma_{\Fr},\ \min(e_{\delta}+e_{\text{rand}},\sqrt{r-r_W})\right\}, \label{lb.eq}
  \end{align}
  where $\vartheta=(\bm X,\bm Z)$, $\mathcal{G}_{\bm W}(\gamma_{\Fr},e_{\delta},e_{\text{rand}})$ is the model class defined in Theorem~S.2, and the infimum is over all measurable estimators $\tilde{\bm U}=\tilde{\bm U}(\hat{\bm X},\bm W)\in\mathcal{O}_{p,r}$.
\end{theorem}

Theorem~\ref{thm:informal_deterministic} gives a three-part decomposition:
  \[
  \underbrace{\gamma_{\Fr}}_{\text{subspace distance}}
  \;+\;
  \underbrace{e_{\delta}}_{\text{mixing bias}}
  \;+\;
  \underbrace{e_{\text{rand}}}_{\text{finite-sample noise}}.
  \]
  The first term $ \gamma_{\Fr} $ is the deterministic distance between $ \mathrm{span}(\bm{U}) $ and $ \mathrm{span}(\bm{W}) $. The second term $ e_{\delta} $ is additional deterministic bias from cross-component mixing, and it becomes larger when mixing is stronger or the gap $D$ is
  smaller. The third term $ e_{\text{rand}} $ is the stochastic error from projected finite-sample perturbation.
  When $r_W=0$, there is no shared knowledge component, so the estimator reduces to the non-transfer baseline estimator $ \hat{\bm{U}}^{(0)} $. In this edge case, the subspace-distance term vanishes ($ \gamma_{\Fr}=0 $), the mixing term is absent, and $e_{\text{rand}}$ is interpreted as the estimation error of $ \hat{\bm{U}}^{(0)} $.

\begin{remark}\label{remark:eigengap_cond}
The condition $D\gtrsim\gamma\delta$ means that the heterogeneous signal $\lambda_{Q_2,\min}$ must be large enough relative to both the misalignment-induced shared contribution $\gamma^2\lambda_{Q_1,\max}$ and the cross-component mixing term $\gamma\delta$.
It can be necessary in mixed-alignment cases with $\delta\neq 0$. In Example~\ref{ex:our}, let $\bm{u}_{q_1}:=\bm{U}\bm{q}_1$, $\bm{u}_{q_2}:=\bm{U}\bm{q}_2$, and $\bm{z}:=\gamma \bm{u}_{q_1}-\sqrt{1-\gamma^2}\,\bm{u}_{\perp}$. Then $\{\bm{u}_{q_2},\bm{z}\}$ is an orthonormal basis for a two-dimensional subspace of $\mathrm{span}(\bm{W}_{\perp})$. In the noiseless setting, the heterogeneous signal direction is recovered from the leading eigendirection of $\bm{W}_{\perp}\bm{W}_{\perp}^{\top}\bm{X}\bm{X}^{\top}\bm{W}_{\perp}\bm{W}_{\perp}^{\top}$, i.e., the projection of $\bm{X}\bm{X}^{\top}$ onto $\mathrm{span}(\bm{W}_{\perp})$. The restriction of this operator to $\mathrm{span}\{\bm{u}_{q_2},\bm{z}\}$ has matrix representation
\[
\begin{bmatrix}
\lambda_{Q_2,\min} & \gamma\delta\\
\gamma\delta & \gamma^2\lambda_{Q_1,\max}
\end{bmatrix}.
\]
Thus $D=\lambda_{Q_2,\min}-\gamma^2\lambda_{Q_1,\max}$ is the diagonal separation, while $\gamma\delta$ is the off-diagonal mixing term. If $D$ does not dominate $\gamma\delta$, the leading eigendirection of this noiseless projected operator cannot stay close to the target heterogeneous direction $\bm{u}_{q_2}$, so a small deterministic error bound is impossible even before sampling noise is added.
When $\delta=0$, the condition reduces to $ \lambda_{Q_2,\min}/\lambda_{Q_1,\max}>\gamma^2 $, the form of a standard population eigengap condition; see, for example, the switch-point gap requirement in \cite{ma2024optimal}.
\end{remark}

The upper bound in Theorem~\ref{thm:informal_deterministic} is the basis for the statistical convergence results in Section~4.2, while the lower bound shows that, over the fixed-$\bm W$ class $\mathcal{G}_{\bm W}(\gamma_{\Fr},e_{\delta},e_{\text{rand}})$, this decomposition is sharp up to constants.

\subsection{Statistical convergence for embedding estimation}

We now derive statistical convergence guarantees for both the concept subspace estimator $ \hat{\bm{U}} $ and the patient embedding estimator $ \hat{\bm{F}} $.
We first establish a subspace error bound, then extend the result to patient embeddings.
\begin{assumption}[Sub-Gaussian error] \label{assum:Z}
    The perturbation matrix $\bm{Z}$ is given by $\bm{Z} = \bm{\Sigma}_{z}^{1/2}\bm{E}$. Here, $\bm{\Sigma}_z$ is a $p\times p$ positive definite matrix with $\|\bm{\Sigma}_z\| \leq \sigma^2$, and $\bm{\Sigma}_{z}^{1/2}$ is its symmetric square root; $\bm{E} = (E_{i,j})_{1\leq i\leq p, 1\leq j \leq n}$ is $p\times n$ with independent sub-Gaussian entries satisfying $\mathbb{E}(E_{i,j}) = 0, \hspace{2mm}\mathbb{E}(E_{i,j}^2) = 1, \hspace{2mm} \|E_{i,j}\|_{\psi_2} \le \phi^2$
    for $1\leq i\leq p, 1\leq j\leq n$, where $\|\cdot\|_{\psi_2}$ is the sub-Gaussian norm and $\sigma^2, \phi^2$ are absolute positive constants.
\end{assumption}
This assumption allows cross-sectional dependence through $\bm{\Sigma}_z$ and is standard in weak-factor analysis \citep{fan2024can,onatski2010determining,onatski2012asymptotics}.

To obtain sharper bounds, we control the heterogeneous basis vectors (the columns of $ \bm{U}\bm{Q}_2 $) individually and then aggregate them. Let $\lambda_{Q_1,1}\ge\cdots\ge\lambda_{Q_1,r_W}$ denote the eigenvalues of $ \bm{Q}_1^{\top}\bm{\Lambda}\bm{Q}_1 $, and let $
\lambda_{Q_2,1}\ge\cdots\ge\lambda_{Q_2,r-r_W}$ denote the eigenvalues of $ \bm{Q}_2^{\top}\bm{\Lambda}\bm{Q}_2 $. Using the rotational freedom in $\bm Q_2$, we choose its basis so that $\bm Q_2^\top\bm\Lambda\bm Q_2=\diag(\lambda_{Q_2,1},\ldots,\lambda_{Q_2,r-r_W})$.

\begin{assumption}[Heterogeneous eigengap] \label{assum:identification_Q}
There exists a fixed constant $c_0\in(0,1)$ such that
$\max\!\left\{\lambda_{Q_2,j+1},\ \gamma^2\lambda_{Q_1,\max}\right\}
\le (1-c_0)\lambda_{Q_2,j}$ for $1\le j\le r-r_W$,
where $\lambda_{Q_2,r-r_W+1}=0$.
\end{assumption}

It is a heterogeneous eigengap condition: each
$\lambda_{Q_2,j}$ must dominate both the next heterogeneous eigenvalue
and the misalignment-induced shared contribution
$\gamma^2\lambda_{Q_1,\max}$. This condition is only needed for the
sharper eigenvector-level bound, not for a coarser block-level bound.

Since our target applications involve rare-disease cohorts with limited sample sizes, we focus on the regime $n\le p$. Let $\lambda_{Q_2,\max}=\max\lambda_{Q_2,j}$ and recall that $\lambda_{Q_2,\min}=\min\lambda_{Q_2,j}, \lambda_{Q_1,\max}=\max\lambda_{Q_1,j}$. Define
\begin{equation}\label{eq:eps_wperp}
  \varepsilon_{W^\perp}^2:=\sum_{j=1}^{r-r_W}\frac{p\lambda_{Q_2,\max}+np}{\lambda_{Q_2,j}^2},
\end{equation}
which captures the stochastic error from projecting sub-Gaussian noise onto $ \mathrm{span}(\bm{W}_{\perp}) $. The following theorem gives a statistical upper bound for subspace estimation.

\begin{theorem}[Subspace estimation error]\label{thm:eigenspace_statistical_ub}
Suppose Assumptions~\ref{assum:Z} and~\ref{assum:identification_Q} hold. If
  \begin{align*}
  \lambda_{Q_2,j}\gtrsim \sqrt{np}+p,\qquad
  D\ge C_\delta \delta,\qquad
  \gamma_{\Fr}\asymp\sqrt{r}\gamma,\qquad
  \varepsilon_{W^\perp}^2\lesssim \varepsilon_0^2,
  \end{align*}
  for a sufficiently large constant $C_\delta=C_\delta(c_0)>0$, then
  \[
  \inf_{\bm{O}\in\mathcal{O}_r}\|\hat{\bm{U}}-\bm{U}\bm{O}\|_{\Fr}^2
  \lesssim
  \min\!\left(
\underbrace{\gamma_{\Fr}^2+\varepsilon_{W^\perp}^2}_{\text{with knowledge}},
  \underbrace{\varepsilon_0^2}_{\text{without knowledge}}
  \right),
  \]
  with probability at least
  $1-C\exp\!\big(-cp+\log((r-r_W)\vee 1)\big).$
  Here $\varepsilon_0$ is the error order of non-transfer baseline estimator $\hat{\bm{U}}^{(0)}$ without auxiliary knowledge.
\end{theorem}

We now briefly interpret the assumptions. First, only the heterogeneous eigenvalues $ \lambda_{Q_2,j} $ in $ \mathrm{span}(\bm{W}_{\perp}) $ need to be lower bounded. The condition $ \lambda_{Q_2,j}\gtrsim\sqrt{np}+p $ is a minimum signal-strength requirement. Compared with classical factor settings that often require signals of order $np$ \citep{bai2002determining,stock2002forecasting}, this allows weaker heterogeneous signals; for example, when $n\asymp p$, it is
satisfied by $ \lambda_{Q_2,j}\gtrsim np^\alpha $ with $0<\alpha<1$ \citep{bai2023approximate,onatski2012asymptotics}. Second, the condition $D \ge C_\delta\delta$ strengthens the $D>2\gamma\delta$ condition in Theorem \ref{thm:informal_deterministic}; see Remark~\ref{remark:eigengap_cond}.
Finally, Remark~\ref{remark:varepsilon_0} shows
that, under standard rates for $ \varepsilon_0 $, the stochastic error $\varepsilon_{W^\perp}$ is typically no larger than $ \varepsilon_0 $.

\begin{remark}[Rate of $\varepsilon_0$]\label{remark:varepsilon_0}
  For fixed $r$, existing results imply $ \varepsilon_{W^\perp}\lesssim\varepsilon_0 $ in standard regimes:
  \begin{itemize}
  \item \textbf{Homogeneous strong signals:} If $\lambda_{Q_2,j}\asymp np$, then $\varepsilon_0=O_p(\sqrt{r/n}+\sqrt{r/p})$
  \citep{bai2002determining,stock2002forecasting}, and $\varepsilon_{W^\perp}\lesssim\varepsilon_0$.

  \item \textbf{Homogeneous weak signals:} If $\lambda_{Q_2,j}\asymp\lambda$, then $\varepsilon_0=O\!\left(\sqrt{rp(\lambda+n)}/\lambda\right)$
  \citep{cai2018rate_optimal}, and $\varepsilon_{W^\perp}\lesssim\varepsilon_0$.

\item \textbf{Heterogeneous weak signals:}
If $\lambda_r\asymp\lambda_{Q_2,\min}\asymp np^\alpha$ with $0<\alpha<1$ and
$\max\{\lambda_1,\lambda_{Q_2,\max}\}\lesssim np$, then, for fixed $r$,
$
\varepsilon_0
=
O_p\!\left(
\sqrt{r}\left(
{p^{1-\alpha}}/{\sqrt{n}}
+
{1}/{p^\alpha}
\right)
\right)
$
\citep[Prop.~6]{bai2023approximate}, and
$\varepsilon_{W^\perp}\lesssim\varepsilon_0$.

  \end{itemize}
  \end{remark}

Theorem~\ref{thm:eigenspace_statistical_ub} gives a no-negative-transfer guarantee for $\hat{\bm{U}}$ by bounding its error with $\min\left(\gamma_{\Fr}^2+\varepsilon_{W^\perp}^2,\ \varepsilon_0^2\right)$.
The knowledge-assisted term has two parts: $ \gamma_{\Fr}^2 $, which captures deterministic misalignment between $ \mathcal{U}=\mathrm{span}(\bm{U}) $ and $
\mathcal{W}=\mathrm{span}(\bm{W}) $, and $ \varepsilon_{W^\perp}^2 $ from \eqref{eq:eps_wperp}, which captures the stochastic error magnitude in $ \mathrm{span}
(\bm{W}_{\perp}) $. Therefore, transfer gains tend to be largest when $ \mathcal{U} $ and $ \mathcal{W} $ overlap strongly (small $ \gamma_{\Fr} $), when fewer signal directions remain in $ \mathrm{span}(\bm{W}_{\perp}) $ (small $r-r_W$), and when heterogeneous eigenvalues $ \lambda_{Q_2,j} $ are large.

\begin{remark}
  \citet{li2024knowledge} explain transfer gains through an enlarged eigengap. In our notation, their Theorem 1 gives
  $O_p\!\left(\gamma_{\Fr}^2 +(\lambda_r^2/\lambda_{Q_2,\min}^2)r/n\right)$.
  When $\gamma_{\Fr}^2$ is non-dominant and $\lambda_{Q_2,\min}\asymp\lambda_r$, this is essentially the same order as standard PCA. Our bound is sharper in heterogeneous spectra because it depends on the full set $\{\lambda_{Q_2,j}\}$ through $\varepsilon_{W^\perp}^2$ rather than only $\lambda_{Q_2, \min}$. Hence, if many $\lambda_{Q_2,j}$ are of order $\lambda_{Q_2,\max}\gg\lambda_r$ (even when some are small), faster rates are still possible. Under Corollary~\ref{coro:simplify_theorem_1}, their rate becomes $O_p(p^{\alpha}/n)$, which is slower than ours.
\end{remark}

\begin{corollary}\label{coro:simplify_theorem_1}
Under Assumption~\ref{assum:Z}, for a fixed transferable $\bm{W}$,
suppose that, for some $0<\alpha<1$, $r\asymp r-r_W\asymp p^\alpha,
p \geq n \gtrsim \log p$,
$\lambda_{Q_2,j}\asymp np^{1+\alpha/2} for
1\le j\le r-r_W-1$,
and
$\lambda_{Q_2,r-r_W}\asymp\lambda_r\asymp np$.
If, for some fixed $c_0\in(0,1)$ and sufficiently large $C>0$,
$
\gamma^2\lambda_{Q_1,\max}
\le (1-c_0)\lambda_{Q_2,r-r_W},
D\ge C\delta,
\gamma_{\Fr}^2\lesssim{p^{0.5\alpha}}/{n},
$
then, with probability at least $1-C_1e^{-c_1p}$, we have
\[
\inf_{\bm{O}\in\mathcal{O}_r}
\|\hat{\bm{U}}-\bm{U}\bm{O}\|_{\Fr}^2
\lesssim
\frac{p^{0.5\alpha}}{n}.
\]
\end{corollary}

We similarly denote by $\hat{\bm{F}}^{(0)}:=\hat{\bm{X}}^\top\hat{\bm{U}}^{(0)} $ the standard non-transfer estimator of $ \bm{F} $, and measure its relative error by $ \varepsilon_{\text{patient}} $:
  \begin{align}\label{eq:epsilon_factor}
  \mathbb{P}\!\left(
  \inf_{\bm{O}\in\mathcal{O}_r}
  \frac{\|\hat{\bm{F}}^{(0)}-\bm{F}\bm{O}\|_{\Fr}}{\|\bm{F}\|_{\Fr}}
  \gtrsim \varepsilon_{\text{patient}}
  \right)
  < \eta_{\varepsilon_{\text{patient}}},
  \end{align}
with $ \varepsilon_{\text{patient}},\eta_{\varepsilon_{\text{patient}}}\to 0 $ as $ n,p\to\infty $. Define
  \begin{equation}\label{eq:eps_F_wperp}
  \varepsilon_{F,W}^2
  :=
  \frac{r_W p }{\lambda_1} + \sum_{j=1}^{r-r_W}
  \frac{p\lambda_{Q_2,\max}+np}{\lambda_{Q_2,j}\lambda_1},
  \end{equation}
where $ \lambda_1 $ is the largest eigenvalue of $ \bm{\Lambda} $ in \eqref{eq:Xhat}.

\begin{theorem}[Patient embedding estimation error]\label{thm:factor_statistical_ub}
  Suppose the conditions in Theorem~\ref{thm:eigenspace_statistical_ub} hold. If $ \varepsilon_0^2+{rp}/{\lambda_1}
\lesssim \varepsilon_{\mathrm{patient}}^2 $, then, with the same probability as in Theorem~\ref{thm:eigenspace_statistical_ub},
  \[
  \inf_{\bm{O}\in\mathcal{O}_r}
  \frac{\|\hat{\bm{F}}-\bm{F}\bm{O}\|_{\Fr}^2}{\|\bm{F}\|_{\Fr}^2}
  \lesssim
  \min\!\left(
  \underbrace{\frac{\lambda_{Q_1,\max}}{\lambda_1}\gamma_{\Fr}^2+\varepsilon_{F,W}^2}_{\text{with knowledge}},
  \underbrace{\varepsilon_{\text{patient}}^2}_{\text{without knowledge}}
  \right).
  \]
\end{theorem}

The bound has the same structure as Theorem~\ref{thm:eigenspace_statistical_ub}: a deterministic term $ (\lambda_{Q_1,\max}/\lambda_1)\gamma_{\Fr}^2 $ and a stochastic term $ \varepsilon_{F,W}^2 $. The first decreases as $ \mathrm{span}(\bm{W}) $ aligns better with $ \mathrm{span}(\bm{U}) $, while
the second comprises the shared-noise
contribution $r_Wp/\lambda_1$ and a heterogeneous component that decreases
as the eigenvalues $\lambda_{Q_2,j}$ grow. 
Using the same argument as in Remark~\ref{remark:varepsilon_0}, one can verify in standard regimes that $ \varepsilon_{F,W}^2\lesssim\varepsilon_{\text{patient}}^2 $. Hence, subspace estimation gains carry over to patient embeddings when $ \gamma_{\Fr} $ is small.

\section{Simulation} \label{sec:simulation}

We generated data from
$\hat{\bm{X}}=\bm{U}\bm{\Lambda}^{1/2}\bm{V}^{\top}+\bm{Z}$,
where $\bm{Z}$ has i.i.d. $N(0,100)$ entries, $\bm{U}=\bm{R}[\bm{I}_r\ \bm{0}_{p-r}]^{\top}$ for a random $\bm{R}\in\mathcal{O}_p$, and $\bm{V}\in\mathcal{O}_{n,r}$. We set $\bm{\Lambda}=\mathrm{diag}(\lambda_1,\dots,\lambda_r)$ with $\lambda_j=np$ for $j\le r-r_W$ and $\lambda_j=\alpha np$ for $j>r-r_W$, where $\alpha\in(0,1)$ controls the strength of the transferable signal relative to the heterogeneous signal.

To mimic imperfect external knowledge, we constructed a raw knowledge matrix $\bm{W}_0=[\bm{W}_{\mathrm{NT}}\ \alpha\bm{W}]\in\mathbb{R}^{p\times d}$, where $\bm{W}\in\mathbb{R}^{p\times r_W}$ is the transferable block and $\bm{W}_{\mathrm{NT}}\in\mathbb{R}^{p\times(d-r_W)}$ contains non-transferable directions. The same $\alpha$ is used in both $\bm{\Lambda}$ and $\bm{W}_0$.

To construct the transferable block $\bm{W}$, we controlled its principal angles with the last $r_W$ columns of $\bm{U}$. For $1\leq j\leq r_W$, we set $\sin^2\theta_j=\bar{\gamma}^2\!\left(1+{(j-1)}/{(r_W-1)}\right)$, where $\bar{\gamma}^2$ controls the principal-angle level. Each knowledge direction is defined as $\bm{w}_j=\cos\theta_j\,\bm{u}_{r-r_W+j}+\sin\theta_j\,\bm{u}^{\perp}_j$,
where $\bm{u}_{r-r_W+j}$ is the $(r-r_W+j)$-th column of $\bm{U}$ and $\bm{u}^{\perp}_j$ is orthogonal to $\mathrm{span}(\bm{U})$. Thus, smaller $\theta_j$ implies stronger alignment with transferable directions. The resulting matrix $\bm{W}=[\bm{w}_1,\dots,\bm{w}_{r_W}]$ has subspace distance
$\gamma_{\Fr}^2=\sum_{j=1}^{r_W}\sin^2\theta_j=1.5\,r_W\,\bar{\gamma}^2$.

Thus, the last $r_W$ latent directions are the transferable components aligned with $\bm{W}$, while the first $r-r_W$ directions remain heterogeneous. We set $\bm{W}_{\mathrm{NT}}$ to have orthonormal columns in the orthogonal complement of $\mathrm{span}(\bm{U})+\mathrm{span}(\bm{W})$.

\subsection{Subspace and patient embedding estimation} \label{subsec:subspace_factor_estimation}

  \begin{figure}[t]
      \centering
      \includegraphics[width=0.85\textwidth]{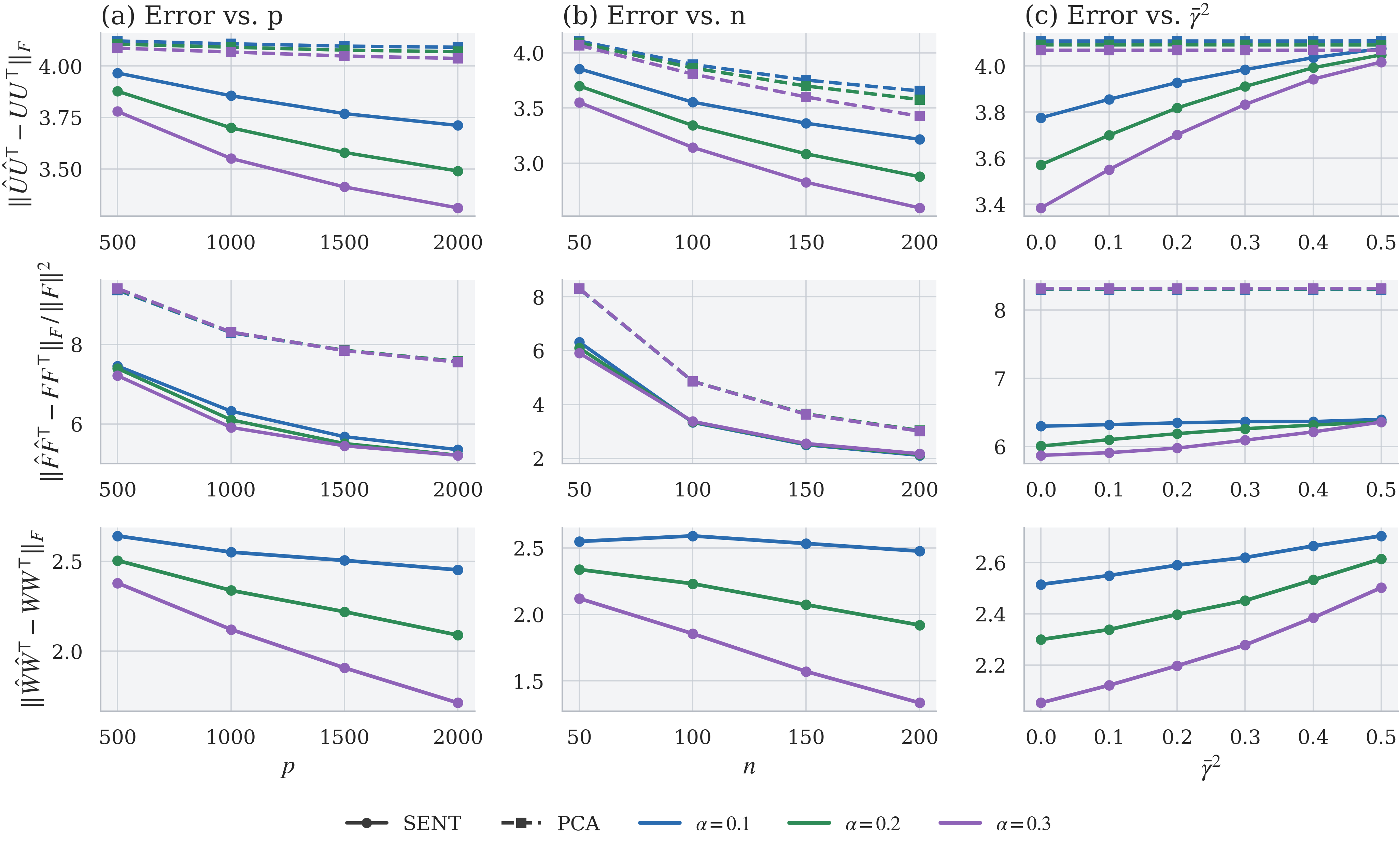}
      \caption{Finite-sample performance of SENT versus PCA under the simulation design in Section~\ref{subsec:subspace_factor_estimation}. Panels vary (a) the number of concepts $p$, (b) the number of patients $n$, and (c) the principal-angle level $\bar{\gamma}^2$. Curves report averages over 100 replicates of the concept subspace error $\|\hat{\bm{U}}\hat{\bm{U}}^{\top}-\bm{U}\bm{U}^{\top}\|_{\Fr}$, the patient embedding error $\|\hat{\bm{F}}\hat{\bm{F}}^{\top}-\bm{F}\bm{F}^{\top}\|_{\Fr}/\|\bm{F}\|^2$, and the knowledge-preprocessing error $\|\hat{\bm{W}}\hat{\bm{W}}^{\top}-\bm{W}\bm{W}^{\top}\|_{\Fr}$. Lower values indicate better recovery.}
      \label{fig:estimation}
  \end{figure}

This subsection evaluates the full SENT pipeline, including knowledge preprocessing and embedding estimation. We focus on the accuracy of estimated concept and patient embeddings, and additionally report how accurately knowledge preprocessing recovers the transferable knowledge matrix.

We varied one factor at a time while fixing the others. Specifically, we varied $p\in\{500,1000,1500,2000\}$ with $n=50$, $\bar{\gamma}^2=0.1$, $r=10$, $r_W=5$, $d=30$, and $\alpha\in\{0.1,0.2,0.3\}$; varied $n\in\{50,100,150,200\}$ with $p=1000$ and the remaining settings unchanged; and varied $\bar{\gamma}^2\in\{0.0,0.1,0.2,0.3,0.4,0.5\}$ with $p=1000$ and $r_W=5$ fixed. Across all settings, $\varepsilon_0$ in the knowledge preprocessing step was estimated by the bootstrap heuristic in Supplementary Section~S2, with $c_W=0.61$.

For each replicate, we report the concept subspace error $\|\hat{\bm{U}}\hat{\bm{U}}^{\top}-\bm{U}\bm{U}^{\top}\|_{\Fr}$, the patient embedding error $\|\hat{\bm{F}}\hat{\bm{F}}^{\top}-\bm{F}\bm{F}^{\top}\|_{\Fr}/\|\bm{F}\|^2$, and the preprocessing error $\|\hat{\bm{W}}\hat{\bm{W}}^{\top}-\bm{W}\bm{W}^{\top}\|_{\Fr}$. Each setting is repeated 100 times, and Figure~\ref{fig:estimation} reports the replicate averages.

Figure~\ref{fig:estimation} shows that both the concept subspace and patient embedding errors decrease as $p$ or $n$ increases and increase with the principal-angle level $\bar{\gamma}^2$. SENT consistently outperforms PCA, especially when $n$ is small or the shared signal is weak. The Step~I preprocessing error generally decreases with $p$, $n$, and $\alpha$, but increases with $\bar{\gamma}^2$. Supplementary Section~S3 further reports transferable rank recovery and an ablation showing that omitting preprocessing can induce negative transfer.

\subsection{Comparison of existing methods} \label{sec:simulation_comparison}

\begin{figure}[t]
    \centering
    \includegraphics[width=1.\textwidth]{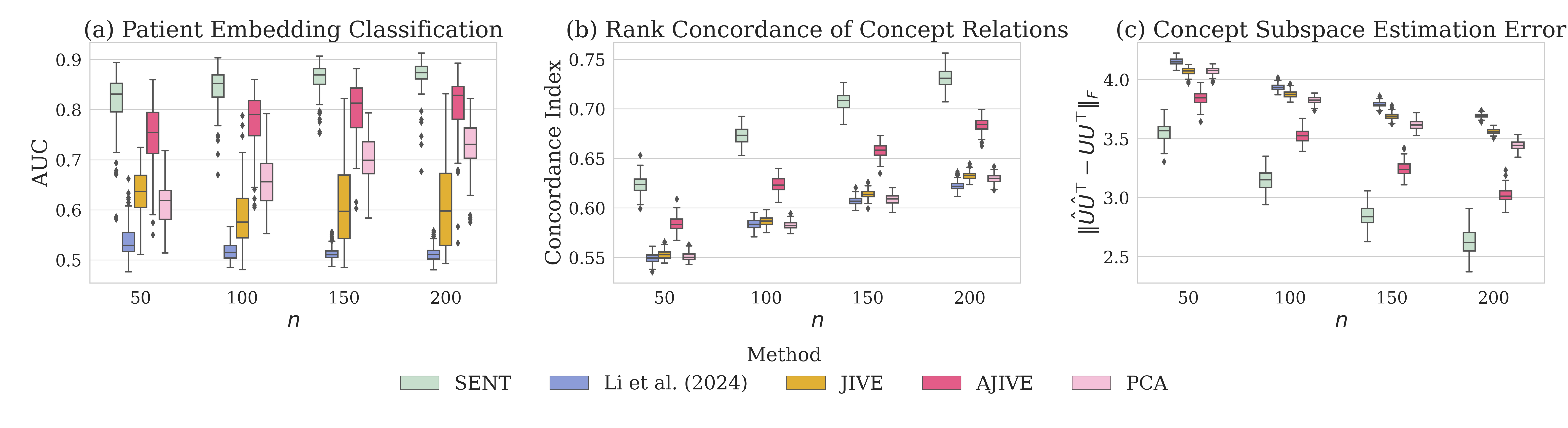}
    \caption{Comparison of SENT and competing methods across sample sizes $n$ under the simulation design in Section~\ref{sec:simulation_comparison}. Panels report (a) area under the receiver operating characteristic curve (AUC) for supervised classification based on estimated patient embeddings, (b) concordance index for estimated concept relations, and (c) concept subspace error $\|\hat{\bm{U}}\hat{\bm{U}}^{\top}-\bm{U}\bm{U}^{\top}\|_{\Fr}$. Higher values are better in (a)--(b), and lower values are better in (c).}
    \label{fig:simulation_comparison}
\end{figure}

In this subsection, we compared SENT with \citet{li2024knowledge}, JIVE \citep{lock2013joint}, AJIVE \citep{feng2018angle}, and PCA under the simulation design in Section~\ref{subsec:subspace_factor_estimation}. We generated $\hat{\bm{X}}\in\mathbb{R}^{p\times n}$ and $\bm{W}_0\in\mathbb{R}^{p\times d}$ as in Section~\ref{subsec:subspace_factor_estimation} and varied only the target sample size, taking $n\in\{50,100,150,200\}$ with $\alpha=0.3$, $p=1000$, $\bar{\gamma}^2=0.1$, $r=10$, $r_W=5$, and $d=30$ fixed. For all methods, we supplied the true rank $r$. SENT was fitted by Algorithms~\ref{alg:data_preprocessing} and~\ref{alg:estimation}, whereas the competing transfer methods were given the true $r_W$ to make the comparison favorable to the baselines.

For concept-relation evaluation, let $\hat{\bm{U}}^{(m)}$ denote the concept subspace estimate from method $m$, let $\hat{\bm{S}}^{(m)}=\hat{\bm{U}}^{(m)}\hat{\bm{U}}^{(m)\top},
\bm{S}=\bm{U}\bm{U}^\top,$
and define $\hat{\bm{s}}^{(m)}=\operatorname{vech}_{\mathrm{off}}(\hat{\bm{S}}^{(m)})$ and $\bm{s}=\operatorname{vech}_{\mathrm{off}}(\bm{S})$, where $\operatorname{vech}_{\mathrm{off}}(\cdot)$ stacks upper-triangular off-diagonal entries. We computed the concordance index between $\hat{\bm{s}}^{(m)}$ and $\bm{s}$; details of the calculation are deferred to Section~S3.

For patient embedding evaluation, we defined a classification task with binary labels $y_i \sim \text{Ber}(\text{sigmoid}(\bm{\beta}^{\top}\bm{f}_i))$, where $\bm{\beta} = (\bm{0}_{r - r_W}^{\top}, \bm{\beta}_1^{\top})^{\top}$ and $\bm{\beta}_1 \in \mathbb{R}^{r_W}$ has i.i.d. $\text{Uniform}(-0.5, 0.5)$ entries. We then generated 1000 additional test patients $\hat{\bm{X}}_{\text{test}}$ from the same model, computed $\hat{\bm{F}}_{\text{test}}^{(m)} = \hat{\bm{X}}_{\text{test}}^{\top}\hat{\bm{U}}^{(m)}$ for each method $m$, trained logistic regression for binary classification on 200 randomly selected test patients, and evaluated AUC on the remainder.

Each experiment was repeated 100 times, and boxplots of AUC (patient embedding classification), concordance index (concept relations), and concept subspace estimation error are shown in Figure~\ref{fig:simulation_comparison}. The patient embedding error is provided in Section~S3. Our SENT method consistently outperforms all baselines across settings, and performance improves with larger $n$. Among baselines, AJIVE outperforms PCA, while JIVE and \citet{li2024knowledge} underperform, likely due to negative transfer.

\section{Real Data Analysis} \label{sec:real_data}

We analyzed EHR data from patients with multiple sclerosis (MS) in the Comprehensive Longitudinal Investigation of MS (CLIMB) registry at Mass General Brigham (MGB). MS is a relatively rare disease, with a reported global prevalence of approximately 35.9 per 100,000 individuals \citep{walton2020rising}, and it exhibits substantial heterogeneity in disease progression and disability outcomes. Quantifying disability severity is central to MS research and clinical management, most commonly through the Expanded Disability Status Scale (EDSS) and the Patient Determined Disease Steps (PDDS) \citep{learmonth2013validation}. In routine clinical data, however, these outcomes are available only for a subset of patients. This makes MS a natural setting for studying whether external knowledge can improve patient and concept representations learned from limited EHR samples.

The observed data matrix $ \hat{\bm{X}}\in\mathbb{R}^{p\times n} $ is a concept-by-patient matrix, where each entry $ \hat{x}_{ij} $ is the observed occurrence count of concept $i$ in patient $j$. Rows correspond to retained clinical concepts (e.g., diagnoses, procedures, medications), and columns correspond to sampled patient windows. To construct the columns, we segmented each patient’s timeline from MS diagnosis into six-month windows, counted concept occurrences within each window, and randomly sampled one window per patient. To stress-test limited-sample regimes, we considered two training sizes, $n=100$ and $500$. For each size, we sampled 100 MS subcohorts (without replacement within each), ran the full estimation and evaluation pipeline on each, and summarized performance across subcohorts.

As external knowledge, we used ONCE embeddings \citep{xiong2023knowledge}, which are derived from large-population EHR corpora and further augmented with contextual biomedical text embeddings. In our framework, these embeddings provide external concept knowledge learned from broader EHR populations. After filtering concepts by ONCE cosine similarity to MS$\geq 0.1$, we formed the raw knowledge matrix $ \bm{W}_0\in\mathbb{R}^{p\times d} $ with $p=4{,}268$ concepts and $d=768$ embedding dimensions.

\begin{figure}[t]
    \centering
    \resizebox{.9\textwidth}{!}{%
        \includegraphics{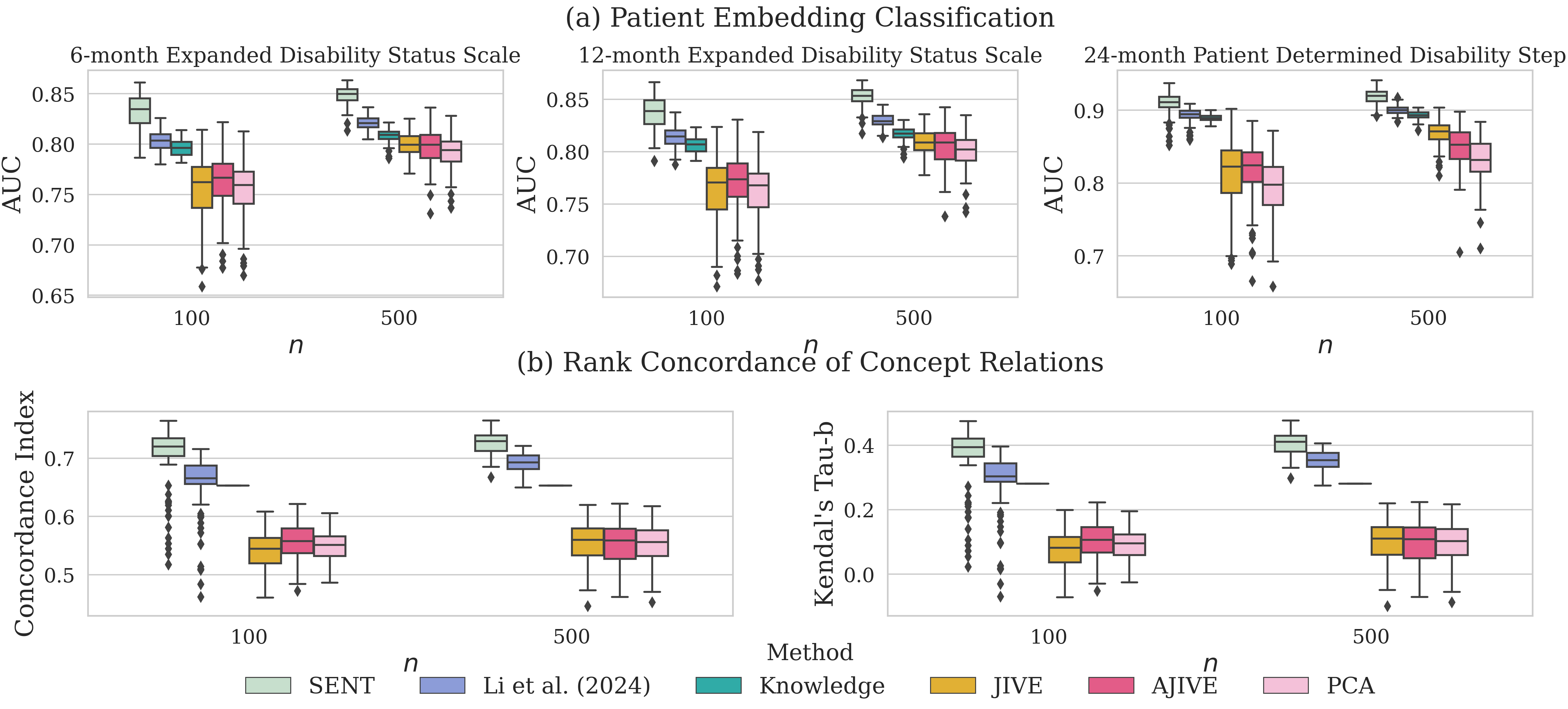}
    }
    \caption{Performance comparison of different methods in the multiple sclerosis (MS) cohort. (a) Area under the receiver operating characteristic curve (AUC) for three supervised classification tasks based on the estimated patient embeddings. (b) Concordance index and Kendall's tau-$b$ statistics between embedding-based MS relation scores and the GPT-based reference ranking.}
    \label{fig:real_temporal_relation}
\end{figure}

We compared SENT with standard PCA, JIVE, AJIVE, \citet{li2024knowledge}, and a direct knowledge-only subspace baseline that uses the external basis without preprocessing (``Knowledge''). Rank $r$ was selected by \textsc{ScreeNOT} in the first repetition and then fixed across repetitions. For SENT, $\varepsilon_0$ was estimated by the subsampling heuristic in Supplementary Section~S2, with $c_W=6$. We selected $r_W$ by \eqref{eq:def_hatr_theta} for SENT and \citet{li2024knowledge}; for JIVE and AJIVE, $r_W$ followed their original procedures.

For concept relation evaluation based on estimated concept embeddings, we formed an evaluation set by taking the union of the top 100 concepts with highest cosine similarity to MS across methods and adding 100 random controls. Let $t$ denote the concept corresponding to multiple sclerosis. For each method $m$, we then quantified the embedding-based relation of concept $i$ to MS by the inner product $s_i^{(m)}=\langle\hat{\bm{u}}_i^{(m)},\hat{\bm{u}}_t^{(m)}\rangle$. We then used GPT-4o \citep{achiam2023gpt} to construct an external reference ranking $g_i\in[0,100]$ for evaluation, and measured rank concordance between $\{s_i^{(m)}\}$ and $\{g_i\}$ using concordance index and Kendall's tau-$b$ correlation coefficient. This evaluation strategy follows \citet{xu2025inference}; the rank concordance formulation, prompt protocol, and robustness analyses are given in Supplementary Section~S4.

\begin{figure}[h!]
    \centering
    \includegraphics[width=1.\textwidth]{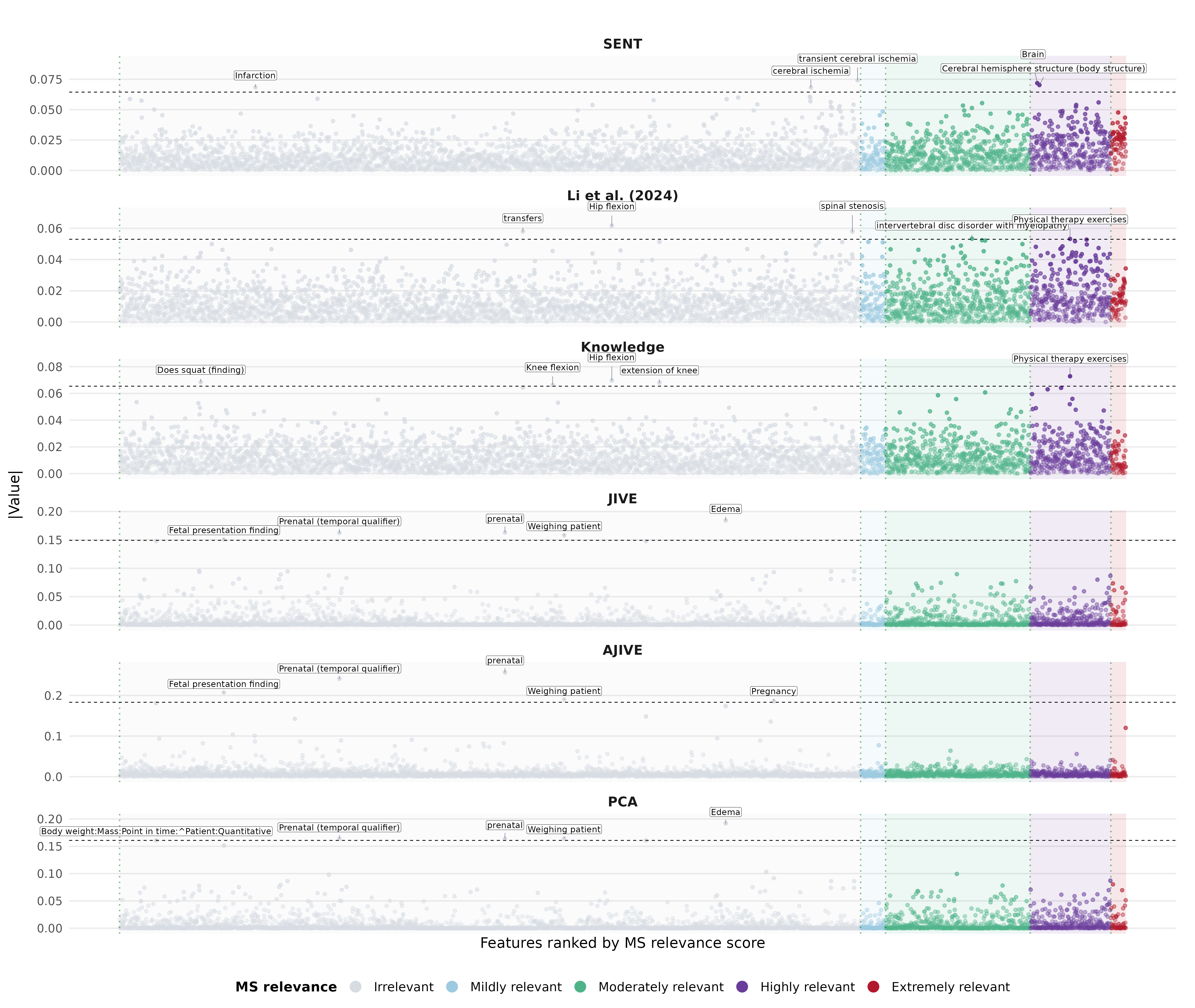}
    \caption{Manhattan-style visualization of the loadings of the most significant factor in the logistic regression for each method. Each point represents a concept and is ranked by GPT-4o relevance scores for multiple sclerosis (MS). Concepts farther to the right are more MS-relevant. We provide textual descriptions of the top five concepts for each method.}
    \label{fig:real_factor_manhattan}
\end{figure}

For patient embedding evaluation, we fixed a held-out set of 1000 patients to form $\hat{\bm{X}}_{\mathrm{test}}$ and computed their patient embeddings using the concept subspace estimated from each training subcohort. We then trained logistic regression on 75\% of the held-out patients using 6-month disability outcomes and evaluated on the remaining 25\%. To assess generalizability across time horizons and outcome sources, we applied the same fitted models to 12-month EDSS and 24-month PDDS outcomes. Supplementary Section~S4 gives the corresponding outcome definitions and preprocessing details.

Figure~\ref{fig:real_temporal_relation} shows that SENT achieves the strongest concept-relation concordance and the best downstream patient-level prediction across all three disability tasks. Among baselines, \citet{li2024knowledge} is typically second, followed by Knowledge; AJIVE gives only modest improvement over PCA, while JIVE is often comparable to PCA.
Figure~\ref{fig:real_factor_manhattan} helps interpret these performance differences. SENT places its largest factor loadings on concepts ranked as highly or extremely relevant to MS, with prominent neurologic and cerebral terms such as brain and cerebral hemisphere structure, while also highlighting some broader cerebrovascular concepts. By contrast, \citet{li2024knowledge} and Knowledge emphasize mobility and rehabilitation-related concepts such as transfers, hip flexion, and physical therapy exercises, whereas PCA, JIVE, and AJIVE place many of their strongest loadings on broad-utilization or nonspecific concepts, including prenatal, pregnancy, weighing patient, body weight, and edema. Overall, these patterns suggest that SENT recovers a representation more concentrated on neurologically meaningful MS-related structure, while weaker baselines are more easily dominated by generic utilization or demographic variation.

\section{Conclusion and Discussion}

SENT addresses a central challenge in rare disease research: learning meaningful patient and concept representations from high-dimensional noisy data when the target cohort is small. Our framework combines knowledge preprocessing with projection-based spectral estimation, allowing transfer from broad-population concept knowledge while reducing the risk of negative transfer from irrelevant directions. Both the simulation study and the MS application show that this preprocessing step is especially valuable when shared structure is weak or the target sample is limited.

More broadly, SENT provides an interpretable and statistically tractable way to incorporate external concept knowledge into unsupervised EHR representation learning. Although we focus on a single knowledge source, the two-step strategy can be extended to multiple knowledge matrices \citep{gu2025robust}. Future work includes combining this framework with nonlinear or semi-supervised spectral methods, and extending it to longitudinal settings where the relationship between data and knowledge may evolve over time.

\paragraph{Data Availability Statement}
The EHR data used in this study come from the Mass General Brigham (MGB) Healthcare System and cannot be shared publicly because of privacy restrictions. The authors do not have authority to release these data.

\paragraph{Acknowledgement.}
The authors thank the editor, the associate editors, and the referees for their constructive comments and suggestions.

\paragraph{Conflict of Interest.} The authors report there are no competing interests to declare. 

{\footnotesize 
\bibliographystyle{apalike}
\bibliography{lfm}}

\end{document}